\documentclass[10pt,journal,compsoc]{IEEEtran}

\usepackage{times}
\usepackage{soul}
\usepackage{url}
\usepackage{array}
\usepackage{graphicx}
\usepackage{subfigure}
\usepackage{amsmath}
\usepackage{amssymb}
\usepackage{amsfonts}
\usepackage[linesnumbered,ruled,vlined]{algorithm2e}
\usepackage{algorithmic}
\usepackage{multirow}
\usepackage{booktabs}
\usepackage{lineno}
\usepackage{pifont}
\usepackage{color}
\usepackage[justification=centering]{caption}
\usepackage{hyperref}

\UseRawInputEncoding
\graphicspath{ {images/} }

\makeatletter
\renewcommand{\@thesubfigure}{\hskip\subfiglabelskip}
\makeatother
\ifCLASSOPTIONcompsoc
  \usepackage[nocompress]{cite}
\else
  \usepackage{cite}
\fi

\ifCLASSINFOpdf

\fi

\begin{document}

\title{Graphfool: Targeted Label Adversarial Attack on Graph Embedding}

\author{Jinyin~Chen, Xiang~Lin, Dunjie~Zhang, Wenrong~Jiang, Guohan~Huang, Hui~Xiong, and Yun Xiang
\IEEEcompsocitemizethanks{

%

\IEEEcompsocthanksitem J. Chen and Yun Xiang are with the Institute of Cyberspace Security, Zhejiang University
of Technology, Hangzhou, 310023, China.

\IEEEcompsocthanksitem W. Jiang is with the Institute of Big Data and Cyber Security, Zhejiang Police College, Hangzhou, 310053, China.

\IEEEcompsocthanksitem J. Chen, X.Lin, D.Zhang, G. Huang, H. Xiong and Y. Xiang are with the College of Information Engineering, Zhejiang University
of Technology, Hangzhou 310023, China.}}




\IEEEtitleabstractindextext{%
\begin{abstract}
Deep learning is effective in graph analysis. It is widely applied in many related areas, such as link prediction, node classification, community detection, and graph classification etc. Graph embedding, which learns low-dimensional representations for vertices or edges in the graph, usually employs deep models to derive the embedding vector. However, these models are vulnerable.  We envision that graph embedding methods based on deep models can be easily attacked using  adversarial examples. Thus, in this paper, we propose Graphfool, a novel targeted label adversarial attack on graph embedding. It can generate adversarial graph to attack graph embedding methods via classifying boundary and gradient information in graph convolutional network (GCN).
Specifically, we perform the following steps: 1),We first estimate the classification boundaries of different classes. 2), We calculate the minimal perturbation matrix to misclassify the attacked vertex according to the target classification boundary. 3), We modify the adjacency matrix according to the maximal absolute value of the disturbance matrix. This process is implemented iteratively. To the best of our knowledge, this is the first  targeted label attack technique. The experiments on real-world graph networks demonstrate that Graphfool can derive better performance than state-of-art techniques. Compared with the second best algorithm, Graphfool can achieve an average improvement of 11.44\% in attack success rate.
\end{abstract}

\begin{IEEEkeywords}
Graph embedding, targeted label attack, node classification, deep learning
\end{IEEEkeywords}}

\maketitle

\IEEEdisplaynontitleabstractindextext

\IEEEpeerreviewmaketitle

\ifCLASSOPTIONcompsoc
\IEEEraisesectionheading{\section{Introduction}\label{sec:introduction}}
\else
\section{Introduction}
\label{sec:introduction}
\fi

\IEEEPARstart{G}{raph} networks are applied in many real-world scenarios, such as social networks~\cite{Borgatti2009Network}, traffic networks~\cite{Latora2002Is}, communication networks~\cite{Kistler2006Cell}, and biological networks~\cite{Montoya2002Small}, etc. The graph embedding, which can learn low-dimensional representations for vertices and edges, provides an effective and efficient way to analyze graphs~\cite{hong2019deep}. It is used in many real-world applications, such as link prediction~\cite{Perozzi2014DeepWalk,Wang2017Signed}, node classification~\cite{Tang2015PTE,Wang2016Linked}, and community detection~\cite{Tian2014Learning,Allab2016A}. By converting a graph into a set of low-dimensional vectors, downstream graph analysis can be more efficient.

Generally, existing graph embedding techniques transform the graph into a similarity graph and calculate its eigenvectors, e.g., IsoMAP~\cite{tenenbaum2000global}, Laplacian eigenmap~\cite{belkin2002laplacian}, and local linear embedding~\cite{roweis2000nonlinear}, etc. As the recent development of machine learning, researchers try to apply the deep learning methods to graph embedding~\cite{goyal2018graph}. Graph neural network (GNN) is a semi-supervised graph embedding method¡£ It utilizes vertex attributes labels to train Ììthe model parameters~\cite{zhou2018graph}, which extends the existing neural network methods into the graph domain~\cite{bronstein2017geometric}. Recently, many GNN models with excellent performance are proposed, such as GCN~\cite{kipf2016semi}, GAE~\cite{kipf2016variational}, and GN~\cite{battaglia2018relational}.

Deep learning techniques are already widely used in security areas, especially in computer vision ~\cite{moosavi2016deepfool, mei2015using,goodfellow2014explaining,biggio2014security,papernot2017practical,elsayed2018adversarial,moosavi2017universal,carlini2017towards,kurakin2018adversarial,shang2017subgraph,faramondi2018network} are catching our attention recently. Adversarial attacks are triggered by carefully crafted adversarial perturbation added to the original image to fool a convolutional neural network (CNN) model~\cite{yuan2019adversarial}. Similarly, in the network area, people care about personal privacy protection against graph analysis~\cite{fard2015neighborhood}. To protect personal privacy against excessive graph mining, individuals require the technique to actively manage their connections and hence, fool the graph analysis tools.

Inspired by adversarial attack in computer vision area~\cite{moosavi2016deepfool}, we propose a novel attack method, Graphfool, on graph embedding. Compared to the existing attack methods, Graphfool can achieve targeted label attack. In essence, we reverse the classification boundary-based optimization procedure of graph convolution network (GCN) model and treat the adjacency matrix as the learning hyperparameters. Our main contributions are summarized as follows:

\setlength{\hangindent}{2em}
$\vcenter{\hbox{\tiny$\bullet$}}$ To the best of our knowledge, this is the first work targeting label adversarial attack on graph embedding. We propose a novel attack method, Graphfool, to invalidate graph embedding algorithms, such as GCN~\cite{kipf2016semi}, DeepWalk~\cite{Perozzi2014DeepWalk}, Node2vec~\cite{grover2016node2vec}, and GraphGAN~\cite{Wang2017GraphGAN}. The experimental results show Graphfool can trick the well-trained deep model with less average number of modified edges(AME) and better attack success rate(ASR).

\setlength{\hangindent}{2em}
$\vcenter{\hbox{\tiny$\bullet$}}$  Unlike the existing graph embedding attack methods, Graphfool can construct targeted label attack by modifying certain edges of the original graph. Therefore, the attacked vertex can be misclassified as any specific class.

\setlength{\hangindent}{2em}
$\vcenter{\hbox{\tiny$\bullet$}}$  Regarding the concealment of Graphfool attack and the processing power of local network, we propose disturbance-limited attack. It controls the perturbation of adjacency matrix to be close to the attacked vertex. Meanwhile, disturbance-limited attack can also reduce the complexity of Graphfool.

The rest of paper is organized as follows. Sec.~\ref{Rel} introduces existing graph attack methods. Sec.~\ref{Preliminaries} discusses the basic theories and techniques of GCN. Sec.~\ref{Method} describes in detail the Graphfool technique. Sec.~\ref{Exp} evaluates our Graphfool method on several real-world data sets. Sec.~\ref{Conclusion} concludes this paper and describes the future works.

\section{RELATED WORK\label{Rel}}
The related work can be generalized into two categories, graph embedding methods and graph attack methods, respectively.

\subsection{Graph embedding methods}
Recently, studies are focused on embedding graph into a low-dimensional vector space based on word2vec model~\cite{mikolov2013efficient}. They are called shallow graph embedding, such as DeepWalk~\cite{Perozzi2014DeepWalk}, Node2vec~\cite{grover2016node2vec}, LINE~\cite{tang2015line}, and GraRep~\cite{cao2015grarep}, etc. DeepWalk~\cite{Perozzi2014DeepWalk} is the first model to learn language from a graph, which uses random walk to sample a sequence for each vertex and treats these generated sequences as sentences using the skip-gram mechanism. Tang et al. propose a novel graph embedding method LINE~\cite{tang2015line}, which is a special case of DeepWalk with the window size of contexts set to one. Inspired by DeepWalk, Grover et al. propose an extension of DeepWalk, called Node2vec~\cite{grover2016node2vec}. Node2vec employs a biased second-order random walk model to provide more flexibility for generating the context vertices. Cao et al. propose an embedding method called GraRep~\cite{cao2015grarep}, which can preserve the node proximities by constructing different k-step probability transition matrices.

In recent years, many deep embedding methods are proposed, which are generally based on deep learning models, such as convolutional neural network (CNN)~\cite{krizhevsky2012imagenet,he2016deep} and generative adversarial networks model (GAN)~\cite{Goodfellow2014Generative}. Kipf et al. propose GCN~\cite{kipf2016semi} for semi-supervised node classification, which scales linearly in the number of graph vertices and learns hidden layer representations by encoding both local graph structure and features of vertices. Similarly, Pham et al. propose column network (CLN), which is a deep learning model for collective classification~\cite{pham2017column}. Compared with GCN, this model emphasises on relation learning, which can process multi-relational data. Monti et al. propose a unified framework MoNet~\cite{monti2017geometric}, which extends the convolution operation to non-Euclidean domains. To strengthen the extraction of critical information, Wang et al. propose non-local neural network~\cite{wang2018non}, which has the ability to capture more detailed information of graph structure. Wang et al. propose GraphGAN~\cite{Wang2017GraphGAN}, which combines two classes of graph representation learning techniques.

\subsection{Graph attack methods}

Adversarial attacks on graph have a wide range of applications. In community detection, Nagaraja propose the first community deception method~\cite{nagaraja2010impact}.  Waniek et al. propose disconnect internally, connect externally (DICE)~\cite{waniek2018hiding}, which ensconces community by randomly deleting edges between members and adding edges between members and non-members.  Fionda et al. propose a novel community deception method based on the safeness~\cite{fionda2018community}. In link prediction, Waniek et al. propose two strategies called closed-triad-removal and open-triad-creation~\cite{waniek2018attack}, which solve the privacy problem caused by link prediction methods. Zhou et al. propose a method to attack local similarity and global similarity by deleting edges~\cite{zhou2018adversarial}. Fard et al. introduce a subgraph perturbation method to randomize the destination of an edge within subgraphs to protect sensitive edges~\cite{fard2012limiting}. They later propose a neighborhood randomization mechanism to probabilistically randomize the destination of an edge within a local neighborhood~\cite{fard2015neighborhood}.

Similar to the adversarial attack in computer vision area, many graph attack techniques are proposed. Dai et al. demonstrate that GNNs are vulnerable by challenging a few edges. They propose reinforcement learning-based method~\cite{dai2018adversarial} to attack graph embedding methods. Z{\"u}gner et al. propose NETTACK, which is an adversarial attack on graph~\cite{zugner2018adversarial}. It generates adversarial graph based on GCN iteratively. Dai et al. propose RL-S2V~\cite{dai2018adversarial}. It can learn to modify the graph structure with the prediction feedback from the target classifier. The modification is implemented by sequentially adding or dropping edges from the graph. Chen et al. propose FGA~\cite{Chen2018Fast} and IGA~\cite{Chen2018Link}. FGA extracts the gradient of pairwise vertices based on the adversarial graph, and then selects the pair of vertices with maximum absolute edge gradient to realize the attack and update the adversarial graph. IGA is designed to mislead the link prediction methods. This method generates adversarial graphs to the target edge. Wang et al. propose Greedy-GAN~\cite{wang2018attack}, which inserts some fake vertices with corresponding fake features into the graph. The attacker is a greedy algorithm, which can generate adjacency and feature matrices of fake vertices. Sun et al. propose Opt-attack~\cite{sun2018data}, which is based on projected gradient descent and attacks unsupervised vertex embedding algorithms, such as DeepWalk and LINE.

\section{Preliminaries\label{Preliminaries}}
First, we define notations which are used throughout the paper. A graph can be represented by  $G=(V,E)$, where $V$ is the vertex set with $\left | V \right |=n$ and $E$ is the edge set. The definition of symbols are listed in the TABLE~\ref{Definition}.

\begin{table}[!ht]
    \centering
    \caption{The definitions of symbols.}
    \label{Definition}
    \begin{tabular}{c|c}
    \hline \hline
    Symbol                        & Definition \\ \hline
    $G=(V,E)$                     &input original network with nodes $V$ and edges $E$ \\
    $\Bar{G}=(V,\Bar{E},M)$       &disturbance network with vertices $V$,edges $\Bar{E}$ and weight $M$\\
    $\Tilde{G}=(V,\Tilde{E})$     &the adjacency matrix of original network $G$\\
    $\Bar{A}$                     &the adjacency matrix added self-connections\\
    $N$                           &the number of vertices of network $G$\\
    $I_N$                         &identity matrix\\
    $Z$                           &the output of the GCN model\\
    $\sigma$                      &$Relu$ active function\\
    $W_i$                         &the weight matrices of GCN model\\
    $X$                           &the feature matrix of all vertices\\
    $C$                           &the number of feature dimensions for each vertex\\
    $H$                           &the number of feature maps for hidden layer\\
    $F$                           &the number of class for vertices in the network\\
    $L$                           &the loss function of the GCN model\\
    $V_L$                         &the set of vertices with labels\\
    $Y$                           &the real label confidence list\\
    $\eta$                        &learning rate\\
    $\theta$                      &the sign function\\
    \hline \hline
    \end{tabular}

\end{table}

\subsection{GCN model}

GCN applies the traditional convolutional neural networks to the graph domain. The model uses an efficient layer-wise propagation rule based on the fist-order approximation of spectral convolutions on graphs. The GCN model can achieved good performance in semi-supervised node classification. In this paper, GCN model is used as the adversarial graph generator to trick other node classification algorithms.

Specifically, we employ a two-layer GCN model with softmax classifier. Its forward model is defined as follows.
\begin{equation}
    Z=softmax(\Bar{A}\sigma(\Bar{A}XW_0)W_1),
\end{equation}
where $X \in R^{N \times C}$ is the feature matrix of all vertices, $\Bar{A}=\Tilde{D}^{-1/2} \Tilde{A}\Tilde{D}^{-1/2}$, $\Tilde{A}=A+I_N$, $\Tilde{D}=\sum_{j}\Tilde{A_{ij}}$, $A$ is the adjacency matrix, and $I_N$ is the identity matrix. Therefore, $\Tilde{A}$ is the adjacency matrix of the graph with self-connections, $\Tilde{D}$ is a degree matrix of $\Tilde{A}$, $W_0\in R^{C\times H}$ is the input-to-hidden weight matrix with the hidden layer of $H$ feature maps, and $W_1\in R^{H\times |F|}$ is the hidden-to-output weight matrix. $\sigma$ denotes the $Relu$ active function. $W_0$ and $W_1$ can be derived using gradient descent training.

The loss function is defined as the cross-entropy error over all labeled examples.
\begin{equation}\label{2}
    L=-\sum_{l=1}^{|V_L|}\sum_{k=1}^{|F|}Y_{lk}\ln(Z_{lk}),
\end{equation}
where $V_L$ is the set of vertices with labels, $|F|$ is the dimension of the output features which is equal to the number of classes, $Z$ is the output feature.

In training process, the GCN model uses the classical gradient descent to optimize the parameters.
\begin{equation}\label{W}
    W_i^{m+1}=W_i^m-\eta \dfrac{\partial L}{\partial W_i^m},
\end{equation}
where $\eta$ is the learning rate. During each iteration, the weights $W_i$, $i\in \left\{ 0,1 \right\}$ are updated.

The GCN model combines vertex features and graph structures using graph convolution. The features of labeled vertices can be mixed with those of unlabeled vertices according to adjacency matrix. Therefore, GCN model can achieve better performance on certain benchmarks~\cite{kipf2016semi}.

\section{The grapthfool method\label{Method}}
Based on two-layered GCN model, Graphfool is designed to achieve targeted label attack by adding or removing few edges of original graph. In this section, we describe  the Graphfool in detail.

\subsection{The framework of graphfool}
\begin{figure}
\centering
    \includegraphics[width=1\linewidth]{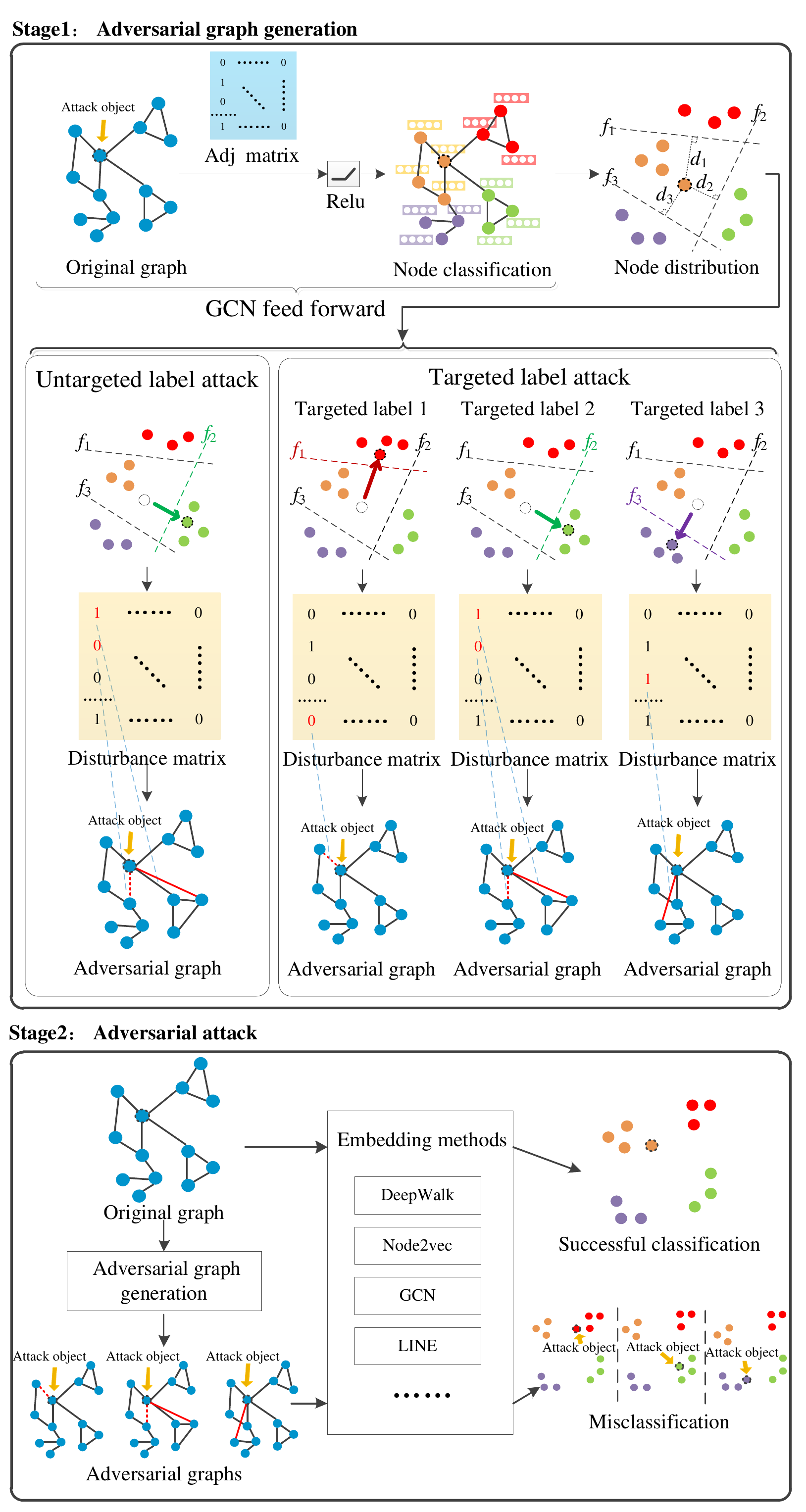}
    \caption{The frame work of Graphfool. }
    \label{framework}
\end{figure}

GCN model with proper training can achieve good performance in node classification. However, if we modify its relationship with other vertices, the node classification result can be completely different. In that case, Graphfool is a adversarial attack technique targeting graph embedding. It consists of two parts, which are adversarial graph generation and adversarial attack, respectively. The framework of Graphfool is shown in Fig.~\ref{framework}.


\setlength{\parskip}{0.5\baselineskip}

\setlength{\hangindent}{2em}
$\vcenter{\hbox{\tiny$\bullet$}}$  \textbf{Adversarial graph generation:} First, we use the original graph and certain labeled vertices to train the GCN model. Then, for each attacked vertex, we use an iterative linearization method to generate minimal adjacency matrix perturbations sufficient to change node classification results. Thus, we derive the final adversarial graph.

\setlength{\parskip}{0\baselineskip}

\setlength{\hangindent}{2em}
$\vcenter{\hbox{\tiny$\bullet$}}$  \textbf{Adversarial attack:} We use the generated adversarial graph to prevent the attacked vertex from the GCN model. Since GCN has excellent generalization ability, the adversarial attack can be implemented on many other graph embedding methods, i.e., the perturbation generated by GCN is universal and the attack has strong transferability.

In adversarial graph generation stage, for a graph dataset, we first derive the node classification results based on adjacency matrix and the trained GCN model. Then for the attacked vertex, we can calculate the classification boundaries ($f_{1}$, $f_{2}$ and $f_{3}$). We can also derive the minimal distance $d_{1}$, $d_{2}$ and $d_{3}$, where $d_{1}>d_{3}>d_{2}$. In untargeted label attack, we select the most vulnerable label whose classification boundary ($f_{2}$) is closest to the attacked node. Then the disturbance matrix is calculated and the adversarial graph can be derived by adding/deleting the edges. For the targeted label attack, it is generally the same as the untargeted label attack. The only difference is that in targeted label attack we should assign the corresponding label of the attacked vertex. In adversarial attack stage, we use the adversarial graph to test the attack effect of Graphfool using various network embedding algorithms.

\subsection{Adversarial graph generator\label{AGG}}
In Sec.~\ref{Preliminaries}, we introduce the structure and training processing of a two-layered GCN. Based on this model, Graphfool generates the adversarial graph.

\subsubsection{Disturbance of adjacent matrix\label{Disturbance}}
For the GCN model, the classifier converts the feature of each vertex to an $F$-dimension vector where $F$ is the number of classes. For each GCN layer, the adjacency matrix is used to capture the structural information of the graph. Therefore, a GCN classifier can be defined as $f:R^{C} \rightarrow R^{F}$. For vertex $i$, its classification is implemented using the following mapping equation.
\begin{equation}\label{6}
    \hat{k}\left ( x_{i},A \right ) = {\mathop{argmax}} f_{k}\left ( x_{i},A \right ),
\end{equation}
where $x_{i}$ is the feature vector of vertex $i$, $A$ is the graph adjacency matrix, $f_{k}\left ( x_{i},A \right )$ is the output of the  $k$-th class of $f\left ( x_{i},A \right )$.

According to Eq.~\ref{6}, if we want to influence the node classification result of vertex $i$, we can change $x_{i}$ and $A$. However, in actual graph networks, such as social networks and communication networks, the feature vectors of vertices are predetermined and thus, difficult to change. However, changing the relationship between the target vertex and other vertices is easier and more concealed. Therefore, Graphfool conducts attacks by changing the adjacency matrix of the graph. In that case, Eq.~\ref{6} can be modified according to the following equation.
\begin{equation}\label{7}
    \hat{k}\left (A \right ) = {\mathop{argmax}} f^{i}_{k}\left (A \right ).
\end{equation}

To simplify the attack problem, we first assume  an linear classifier $f^{i}\left (A \right )$, i.e., $f^{i}\left (A \right )=W^TA+b$, where $W$ and $b$ are trained classifier parameters. The goal is to make the target vertex misclassified by adding minimal perturbation on $A$. Therefore, the attack problem can be modeled as follows.
\begin{equation}
\begin{matrix}
    {\mathop{argmin}}{\left \| R \right \|}_{2} \\
    s.t.\exists k: {w_{k}}^{T}(A_{0}+R)+b_{k}\geq {w_{\hat{k}(A_{0})}}^{T}(A_{0}+R)+b_{\hat{k}(A_{0})},
\end{matrix}
\label{8}
\end{equation}
where $A_{0}$ is the original adjacency matrix of the graph, $R$ is perturbation matrix, and $w_{k}$ is the $k$-th column of $W$. Thus, for the $k$-th class, we can construct class boundary function between $f^{i}_{k}\left (A \right )$ and $\hat{k}\left (A \right )$ using the following equation.
\begin{equation}
    f^{i}_{k}\left (A \right )-\hat{k}\left (A \right ) = 0 (k \neq \hat{k}\left (A_{0} \right )).
\label{9}
\end{equation}

For original adjacency matrix $A_{0}$, the minimal perturbation $R(A_{0})$ corresponds to the minimal distance between $A_{0}$ to these boundaries. It can be calculated using the following equation.

\begin{small}
\begin{equation}
    R(A_{0})= \mathop{argmin}_{k\neq \hat{k}(A_0)}\dfrac{|f^{i}_{k}\left (A_{0} \right )-\hat{k}\left (A_{0} \right )|(w_{k}^{T}(A_{0})-w_{\hat{k}(A_{0})}^{T}(A_{0}))}{\lVert w_{k}^{T}(A_{0})-w_{\hat{k}(A_{0})}^{T}(A_{0})\rVert^2_2}.
\label{10}
\end{equation}
\end{small}

For general non-linear classifiers, we approximate the node classification boundary function using the first-order Taylor expansion of each classifier.
\begin{small}
\begin{equation}
    f^{i}_{k}\left (A_{0} \right )-\hat{k}\left (A_{0} \right ) + \triangledown{{f^{i}_{k}\left (A_{0} \right )}^{T}}A - \triangledown{{\hat{k}\left (A_{0} \right )}^{T}}A= 0 (k \neq \hat{k}\left (A_{0} \right )).
\label{11}
\end{equation}
\end{small}
$R(A_{0})$ can be derived similarly using Eq.~\ref{10}.

To guide the change of $A_{0}$, we need to calculate the value of $R(A_{0})$. Since the adjacent matrix of an undirected graph is symmetric, we symmetrize  $R(A_{0})$ to obtain $\hat{R}(A_{0})$ as shown in the following equation.

\begin{equation}
    \hat{R}_{ij}(A_{0}) = \hat{R}_{ji}(A_{0}) = \left\{\begin{matrix}
\frac{R_{ij}(A_{0}) + R_{ji}(A_{0})}{2} &i\neq j\\
0 & i=j.
\end{matrix}\right.
\label{12}
\end{equation}

In Eq.~\ref{12}, the elements in $\hat{R}(A_{0})$ have continuous values. For a specific element $\hat{R}_{ij}(A_{0})$ in $\hat{R}(A_{0})$, its positive/negative value  indicates that we should adding/deleting the edge between the pair of vertices $(v_{i},v_{j})$. The larger value of $\left | \hat{R}_{ij}(A_{0}) \right |$ indicates the added/deleted edge can influence the classification result of the target node more significantly.

\subsubsection{Adversarial graph generator}

In this section, we propose an adversarial graph generator based on our adjacent matrix disturbance generation technique. We modify one edge during each iteration and the generation process runs for $K$ iterations. To avoid excessive perturbation, the total iteration number $K$ should be limited. For a given graph, its average degree represents the sparseness of the graph. Therefore, we limit $K$ based on the graph average degree. The generation process is iterative.

The flows of the generation process is shown in Alg. 1.

\begin{enumerate}
\item We construct $\hat{R}(A_{h-1})$ based on $A_{h-1}$. Using Eq.~\ref{9} and Eq.~\ref{10}, we calculate the classification surface closest to $A_{h-1}$, its target label $l_{h-1}$, and the perturbation $\hat{R}(A_{h-1})$  with $A_{0}=A$.
\item  We select perturbation edges. Based on $\hat{R}(A_{h-1})$, we select a pair of vertices ${(v_i, v_j)}$ which has maximal absolute value $\hat{R}_{ij}(A_{h-1})$. It should be noted that if $\hat{R}_{ij}(A_{h-1})$ is positive/negative and $v_i$ and $v_j$ are connected/disconnected in $A_{h-1}$, we cannot further add or delete the edge between this pair of vertices. Hence, we just ignore such pairs of vertices in the process and continue.
\item  We update the adjacency matrix $A_{h}$. We modify the $(h-1)$-th adjacency matrix with selected pair of vertices and generate a new adversarial graph. The $h$-th adjacency matrix is calculated using the following equation.
\begin{equation}
    A_{h\_ij} = A_{h-1\_ij} + \theta(\hat{R}_{ij}(A_{h-1}))
\label{13}
\end{equation}
where $A_{h\_ij}$ and $A_{h-1\_ij}$ are the elements of $A_{h}$ and $A_{h-1}$ and $\theta(\hat{R}_{ij}(A_{h-1}))$ is the signed value of the pair of elements with maximal absolute values in perturbation matrix $\hat{R}(A_{h-1})$.
\end{enumerate}

\begin{algorithm}
\caption{Adversarial graph generator via GCN}
\KwIn{Original graph $G$, number of iteration $K$.}
\KwOut{The adversarial graph $\Tilde{G}$.}
 Train the GCN model on original graph $G$ to obtain model parameters $W$ via Eq.~\ref{W}.\\
 Initialize the adjacency matrix of the adversarial graph by $A_{0}=A$;\\
 \For{$h=1$ to $K$}
{    Construct $\hat{R}(A_{h-1})$ based on $A_{h-1}$;\\
     Select the perturbation edges which has maximum absolute value in $\hat{R}(A_{h-1})$;\\
     Update the adjacency matrix $A_{h}$ according to \\
     $A_{h\_ij} = A_{h-1\_ij} + \theta(\hat{R}_{ij}(A_{h-1}));$
        }
\textbf{Return} the adversarial graph $\Tilde{G}$, with the adversarial adjacency matrix $A_{K}$\\
\label{algorithm1}
\end{algorithm}

\subsection{Targeted label attack}

In the previous sections, we introduce our technique to generate adversarial graph based on  classification boundary and minimal disturbance of adjacency matrix. Moreover, Graphfool can also perform targeted label attack. Thus, we define a new goal, i.e., for vertex $i$, we want to misclassify it into class $l(l \neq \hat{k}\left (A_{0} \right ))$. Thus, the attack problem can be generalized as follow.
\begin{equation}
\begin{matrix}
    {\mathop{argmin}}{\left \| R \right \|}_{2} \\
    s.t.\forall k(k \neq l): {w_{l}}^{T}(A_{0}+R)+b_{l}\geq {w_{k}}^{T}(A_{0}+R)+b_{k}
\end{matrix}.
\label{14}
\end{equation}

In general, during iteration $h$, when $f^{i}_{l}\left (A_{h-1} \right ) \neq \hat{k}\left (A_{h-1} \right )$, the goal is to add the minimal perturbation $R$ to make vertex $i$ crossover the classification boundary $f^{i}_{l}\left (A \right )-\hat{k}\left (A \right ) = 0$. This minimum perturbation can be derived using the following equation.
\begin{small}
\begin{equation}
    R(A_{h-1})= \dfrac{|f^{i}_{l}\left (A_{h-1} \right )-\hat{k}\left (A_{h-1} \right )|(w_{l}^{T}(A_{h-1})-w_{\hat{k}(A_{h-1})}^{T}(A_{h-1}))}{\lVert w_{l}^{T}(A_{h-1})-w_{\hat{k}(A_{h-1})}^{T}(A_{h-1})\rVert^2_2}.
\label{15}
\end{equation}
\end{small}

\subsection{Transferring adversarial attack}
Besides the baseline GCN model, we can also use the modified adjacency matrix to attack other node classification methods. Most node classification algorithms rely upon the connection relationship between vertices. Vertices with strong relationship are typically divided into the same class. Therefore, these algorithms have similar decision boundaries. In that case, the GCN based adversarial attack can also be effective on many other node classification methods. In the experiment section, we use the adversarial graph generated by Graphfool to attack other node classification algorithms. The experimental results show strong transferability of our technique.

\section{Experimental results\label{Exp}}
To validate our technique, we test it for both untargeted label attack and targeted label attack. Moreover, to demonstrate the concealment of Graphfool, we also perform the single-edge attack and disturbance-limited attack.
\subsection{Experimental setup}
Our experiments are performed on a machine with i7-7700K 3.5GHzx8 (CPU), TITAN Xp 12GiB (GPU), 16GBx4 memory (DDR4), and Ubuntu 16.04 (OS).
\subsubsection{Datasets}
In the experiment, we test different techniques on three datasets, which are Cora, Citeseer, and Pol.blogs, respectively. Their statistics are provided in TABLE~\ref{dataset_table}.
\begin{table}[!ht]
    \centering
    \caption{The basic statistics of three network datasets.}
    \label{Datasets}
    \begin{tabular}{c|c|c|c}
    \hline \hline
    Datasets&Nodes&Links&Classes  \\
    \hline
         Cora&2708&5429&7 \\
         Citeseer&3312&4732&6 \\
         Pol.blogs&1490&19090&2\\
    \hline \hline
    \end{tabular}
    \label{tab:my_label}
\label{dataset_table}
\end{table}

\begin{table*}[!t]
\centering
\caption{The attack effects obtained by different attack methods toward varous network embedding methods on multiple datasets.}
\label{random_result}
\resizebox{\linewidth}{!}{
\begin{tabular}{c|c|ccc|ccccc|ccc|ccccc}
\hline \hline
\multicolumn{1}{c|}{\multirow{3}{*}{Datasets}}&
\multicolumn{1}{c|}{\multirow{3}{*}{Model}}&\multicolumn{8}{c|}{ASR(\%)}& \multicolumn{8}{c}{AME}\\
\cline{3-18}
\multicolumn{1}{c|}{}& \multicolumn{1}{c|}{} & \multicolumn{3}{c|}{Graphfool}& \multicolumn{5}{c|}{BASELINE}& \multicolumn{3}{c|}{Graphfool}&\multicolumn{5}{c}{BASELINE} \\
\cline{3-18}

\multicolumn{1}{c|}{} & \multicolumn{1}{c|}{}&unlimited&direct&indirect&FGA&NATTECK&RL-S2V&GradArgmax&\multicolumn{1}{c|}{DICE}& unlimited&direct&indirect&FGA&NATTECK&RL-S2V&GradArgmax&\multicolumn{1}{c}{DICE}\\ \hline
\multirow{5}{*}{Cora} & GCN &\textbf{100}&100&89.66&100&92.87&93.83&90.32&54.95&\textbf{1.78}&1.92&5.60&2.54&6.09&6.65&7.02&9.13 \\
&GraRep &\textbf{100}&100&87.53&100&97.22&100&95.35&89.09&\textbf{5.43}&5.57&9.41&5.56&5.94&5.96&7.29&7.37 \\
&Deepwalk &\textbf{100}&100&76.70&100&94.06&95.40&90.95&93.52&5.57&\textbf{5.13}&11.07&5.61&7.24&6.92&7.89&7.20 \\
&node2vec &\textbf{100}&100&77.57&100&97.29&100&96.24&89.09&\textbf{4.94}&4.94&10.21&5.66&6.75&6.14&7.44&7.37 \\
&LINE &\textbf{100}&100&85.72&100&96.34&95.51&89.98&88.99&\textbf{5.38}&5.47&9.41&5.64&7.02&6.96&8.10&7.66 \\
&GraphGAN &\textbf{100}&100&74.77&100&92.26&95.56&88.24&84.55&5.63&\textbf{5.05}&10.65&5.65&8.82&6.90&8.02&7.96 \\
\cline{2-18}
&Average &\textbf{100}&100&81.99&100&95.01&96.72&91.85&83.37&4.79&\textbf{4.68}&9.39&5.11&6.98&6.59&7.63&7.78 \\
\hline
\multirow{5}{*}{Citeseer} & GCN &\textbf{100}&100&93.55&100&87.50&91.84&88.33&70.37&\textbf{1.42}&1.32&4.65&3.52&6.88&5.86&6.62&9.87 \\
&GraRep &\textbf{100}&100&97.63&100&94.28&94.44&89.23&93.22&\textbf{5.27}&5.89&8.25&5.32&6.51&6.94&6.89&7.56 \\
&Deepwalk &\textbf{100}&100&81.03&100&96.96&94.34&90.96&93.44&\textbf{5.38}&5.74&5.38&5.68&7.06&6.56&6.90&7.08 \\
&node2vec &\textbf{100}&100&79.31&100&93.93&93.88&89.09&91.38&\textbf{4.48}&4.75&9.33&5.62&6.34&7.02&6.80&7.13 \\
&LINE &\textbf{100}&100&98.36&100&95.82&93.88&86.66&96.72&\textbf{5.76}&6.25&7.95&5.88&6.02&6.80&7.26&7.21 \\
&GraphGAN &\textbf{100}&100&77.19&100&92.06&94.12&85.32&88.24&\textbf{5.54}&5.58&9.25&5.91&7.42&7.04&7.79&8.26 \\
\cline{2-18}
&Average &\textbf{100}&100&87.85&100&93.43&93.75&88.27&88.90&\textbf{4.64}&4.92&5.16&7.46&6.71&6.70&7.04&7.98 \\
\hline
\multirow{5}{*}{Pol.blogs} & GCN &\textbf{95.25}&93.62&21.27&87.87&82.97&82.98&78.34&50.27&\textbf{4.92}&5.91&17.49&8.42&11.89&9.09&10.21&11.85 \\
&GraRep &\textbf{94.87}&94.87&5.28&83.88&79.91&79.17&75.69&61.06&\textbf{7.21}&7.56&19.36&9.58&10.48&11.02&11.88&14.22 \\
&Deepwalk &95.25&\textbf{97.87}&4.26&84.26&75.41&78.72&76.25&64.52&6.26&\textbf{6.01}&18.96&9.84&10.06&10.09&11.04&12.35 \\
&node2vec &\textbf{97.87}&97.87&6.38&84.34&78.32&79.17&73.32&67.89&7.13&\textbf{7.04}&18.72&9.72&10.58&10.89&11.62&14.86 \\
&LINE &\textbf{96.52}&95.87&5.47&85.25&76.35&75.00&70.26&66.74&\textbf{6.84}&7.12&19.56&9.90&10.26&11.30&11.73&12.82 \\
&GraphGAN &\textbf{95.74}&95.74&8.51&81.21&72.26&79.17&72.02&64.58&\textbf{7.61}&7.85&18.85&9.41&11.08&11.55&12.04&12.26 \\
\cline{2-18}
&Average &95.91&95.97&8.52&84.47&77.54&79.02&74.31&62.51&6.66&6.92&18.82&9.48&10.73&10.66&11.42&13.06 \\
\hline
\hline
\end{tabular}}
\label{T3}
\end{table*}

\setlength{\parskip}{0.5\baselineskip}

\setlength{\hangindent}{2em}
$\vcenter{\hbox{\tiny$\bullet$}}$ \textbf{Cora:} The Cora dataset consists of 2708 scientific publications categorized into seven classes~\cite{Mccallum2000Automating}. The citation graph consists of 5429 edges. Each edge represents the citation relationship.
\setlength{\parskip}{0\baselineskip}

$\vcenter{\hbox{\tiny$\bullet$}}$ \textbf{Citeseer:} The Citeseer dataset consists of 3312 scientific publications categorized into six classes~\cite{Mccallum2000Automating}. The citation network consists of 4732 edges.

$\vcenter{\hbox{\tiny$\bullet$}}$ \textbf{Pol.blogs:} The Pol.blogs dataset~\cite{adamic2005political} shows the political divide of blog. It contains 1490 vertices and 19090 edges. The vertices are divided into two classes.

\subsubsection{Evaluation criteria}
In this section, we introduce the evaluation criteria for the comparing methods.
\begin{itemize}
\item Attack success rate (ASR): We attack a series of vertices in the graph and the ASR is defined as follows.
\begin{equation}
    ASR=\frac{N_s}{N}\times100\%,
\end{equation}
where $N_s$ is the number of misclassified vertices and $N$ is the total number of nodes being attacked.
\item The average number of modified edges (AME): To attack the target vertex, we add/delete edges between vertices. The modifications should be minor and undetectable. Thus, the method with smaller AME is better. The AME equation is defined as follows.
\begin{equation}
    AME=\frac{1}{N}\sum_{i=1}^N L_i,
\end{equation}
where $N$ is the total number of nodes being attacked and $L_i$ is the number of modified edges for vertex $i$.
\end{itemize}

\subsubsection{Comparing methods}
To validate our Graphfool technique, we compre it with five state-of-art graph embedding attack techniques shown as follows.
\begin{itemize}
\item \textbf{DICE}~\cite{waniek2018hiding}: For each iteration, DICE removes $b$ edges of the target vertex randomly and then add edges between the target vertex and $K-b$ vertices of different classes.
\item \textbf{NETTACK}~\cite{zugner2018adversarial}: NETTACK selects key edges based on pivotal data characteristics, e.g., degree distribution. Then it uses two scoring functions to measure the change in the confidence value. After modifying an edge, it uses the feature with highest score to update the adversarial graph according to the confidence value change.
\item \textbf{FGA}~\cite{Chen2018Fast}: FGA extracts the gradient of pairwise vertices based on the original graph. Then it selects the pair of vertices with the maximal absolute edge gradient to update the adversarial graph.
\item \textbf{RL-S2V }~\cite{dai2018adversarial}: RL-S2V is a hierarchical reinforcement learning based attack method. It learns a Q-function parameterized by S2V to perform a generalized attack.
\item \textbf{GraArgmax}~\cite{dai2018adversarial}: GradArgmax calculates the gradient of adjacency matrix based on the output of each hidden layers and the loss function. Then it adopts greedy algorithm to select the pairs of vertices to attack the original graph.
\end{itemize}

\subsection{Evaluation results}
In this section, we evaluate our Graphfool method on three real-world datasets. The techniques are tested with untargeted label attack, targeted label attack, single-edge attack, and disturbance-limited attack.

\subsubsection{Untargeted label attack}

First, we randomly select 20 vertices in each category to form the set of attacked vertices. To analyze the relationship between modified edges and attacked vertices, we consider direct, indirect, and unlimited attacks, respectively~\cite{Chen2018Fast}.
\begin{itemize}
\item \textbf{Direct attack:} This attack method only attacks the edges directly connected to the attacked vertex.
\item \textbf{Indirect attack:} This attack method attacks the edges not directly connected to the attacked vertex.
\item \textbf{Unlimited attack:} This attack method can remove or add edges between any pair of vertices.
\end{itemize}

Without loss of generality, we assume the number of modified edges less than 20 for each attack. The attack results are shown in TABLE~\ref{T3}. For the unlimited attack case, Graphfool outperforms the other attack methods in most of the cases, in terms of higher ASR and lower AML. In Cora and Citeseer, for the unlimited case, both Graphfool and FGA achieve 100\% ASR. However, Graphfool has significantly smaller AME, which implies that the adversarial graph generated by Graphfool has much less perturbation. For Pol.blogs, our technique can get 96.03\% ASR and 6.48 AME. The performance deteriorates a bit because of the denseness of the network (with average degree close to 25.6). However, our techinique still outperform other any other comparing techniques.

For our Graphfool technique, the unlimited and direct attacks have relatively close performance. However, the indirect attack has relatively worse performance.  This demonstrates that direct attacks are typically more effective than indirect ones.  In graphfool, we generate a adversarial graph through the GCN model. The generated adversarial graph is then used as input to attack other node classification algorithms. TABLE~\ref{T3} shows the attacking results. Although the adversarial graph is generated based on GCN model, it can also achieve excellent attack performance for other node classification algorithms. This demonstrated the strong transferability of our technique. Moreover, our adversarial graph achieves less AME on GCN model. This phenomenon indicates that the adversarial graph can capture the vulnerability of GCN model more accurately.

Moreover, for the datasets of Cora and Citeseer, where the graphs are relatively sparse, indirect attacks can achieve relatively high attack performance, which is similar to DICE of direct attack. This implies that we may be able to change the edges far away from the target nodes to perform the attack. In other words, the local structure of these vertices is not necessarily destroyed, making the attack harder to detect. For the dataset of Pol.Blogs, since it is very dense, the performance of graphfool is limited, especially for indirect attack.

\subsubsection{Targeted label attack}

In this section, we perform targeted label attack for the Cora and Citeseer dataset, which have more than 2 labels. For each dataset, we also randomly select 20 vertices in each category to form the set of attacked vertices. The specific attack strategy of Graphfool is the unlimited attack.

The results of the targeted label attack are shown in TABLE~\ref{T4} and TABLE~\ref{T5}. Compared with untargeted label attack, the ASR of targeted label attack has a significant decline, while its AME also increases. It implies that the targeted label attack is not as effective as untargeted label attack. The reason is that due to the explicit directionality of targeted label attack, its generated disturbance has more conditional constraints. These constraints can increase the cost of ASR and AME. \textbf{}{***(what does these sentences mean? Your method has random performance on different labels?? Why??)In addition, in the Cora dataset, when the targeted label is 3, the method of targeted label attack could achieve higher ASR and lower AME, which means that other vertices are more likely to disguise as the vertices with label 3. And the same phenomenon also appears on Citeseer dataset. We can conclude that the vertices with different labels have different properties, which leads to different ASR and AML under different targeted labels on targeted label attack.}

\begin{table}[!t]
\centering
\caption{Targeted label attack on Cora.}
\resizebox{\linewidth}{!}{
\begin{tabular}{c|c|ccccccc}
\hline \hline
\multirow{2}{*}{Metris} & \multirow{2}{*}{Model} & \multicolumn{7}{c}{Targeted Label} \\ \cline{3-9}
                        &                      & 0   & 1   & 2   & 3   & 4   & 5   & 6  \\ \hline
\multirow{7}{*}{ASR(\%)}& GCN                  &73.96&70.79&66.67&98.86&84.27&79.12&66.30\\
                        & GraRep               &72.92&68.54&67.82&96.59&82.02&79.12&65.22\\
                        & DeepWalk             &73.96&67.42&66.67&95.45&84.27&78.02&65.22\\
                        & Node2vec             &72.92&70.79&65.52&94.32&83.15&76.92&61.53\\
                        & LINE                 &70.83&68.54&65.52&96.59&85.39&76.92&64.13\\
                        & GraphGAN             &72.92&71.91&66.67&94.32&85.39&79.12&61.53\\ \cline{2-9}
                        & Average              &72.92&69.67&66.48&96.02&84.08&78.20&63.99\\ \hline
\multirow{7}{*}{AME}    & GCN                  &6.28 &6.91 &7.56 &2.00 &4.57 &5.47 &8.03\\
                        & GraRep               &7.03 &7.41 &7.58 &3.25 &5.03 &6.31 &8.46\\
                        & DeepWalk             &6.95 &7.28 &7.69 &3.14 &5.08 &6.16 &8.53\\
                        & Node2vec             &7.03 &7.37 &7.63 &3.26 &5.11 &6.28 &8.23\\
                        & LINE                 &7.26 &7.37 &7.73 &3.23 &4.93 &6.34 &8.37\\
                        & GraphGAN             &6.96 &7.45 &7.60 &2.99 &4.72 &6.42 &8.23\\ \cline{2-9}
                        & Average              &6.75 &7.30 &7.63 &2.98 &4.91 &6.16 &8.31\\ \hline \hline
\end{tabular}}
\label{T4}
\end{table}

\begin{table}[!t]
\centering
\caption{Targeted label attack on Citeseer.}
\resizebox{\linewidth}{!}{
\begin{tabular}{c|c|cccccc}
\hline \hline
\multirow{2}{*}{Metris} & \multirow{2}{*}{Model} & \multicolumn{6}{c}{Targeted Label} \\ \cline{3-8}
                        &                      & 0   & 1   & 2   & 3   & 4   & 5    \\ \hline
\multirow{7}{*}{ASR(\%)}& GCN                  &83.78&78.46&74.19&92.19&85.16&88.52  \\
                        & GraRep               &85.14&78.46&79.03&90.63&83.33&90.16  \\
                        & DeepWalk             &82.43&78.46&80.65&90.63&87.04&86.89  \\
                        & Node2vec             &85.14&76.92&77.41&89.06&81.48&85.25  \\
                        & LINE                 &82.43&80.00&75.81&93.75&87.04&86.89  \\
                        & GraphGAN             &81.08&76.92&72.58&89.06&81.48&83.60  \\ \cline{2-8}
                        & Average              &83.33&78.20&76.61&90.89&84.26&86.89  \\ \hline
\multirow{7}{*}{AME}    & GCN                  &4.55 &5.72 &5.56 &3.11 &4.39 &3.77   \\
                        & GraRep               &6.44 &7.13 &6.92 &4.78 &5.59 &4.33   \\
                        & DeepWalk             &6.64 &6.92 &6.69 &4.63 &5.61 &4.30   \\
                        & Node2vec             &6.70 &7.02 &6.95 &4.46 &5.76 &4.33   \\
                        & LINE                 &6.64 &6.88 &6.71 &4.48 &5.63 &4.18   \\
                        & GraphGAN             &6.53 &7.12 &6.92 &4.28 &5.69 &4.26   \\ \cline{2-8}
                        & Average              &6.25 &6.80 &6.62 &4.29 &5.45 &4.20   \\ \hline \hline
\end{tabular}}
\label{T5}
\end{table}

\subsubsection{Single-edge attack}

In computer vision area, other than the success rate of adversarial attack, minimizing its disturbance is also an important goal~\cite{moosavi2016deepfool}. Similarly, in graph based attacks, if an attack method could get close performance with fewer modified edges, it has a better attack concealment. In this section, to evaluate the concealment of Graphfool, we design a single-edge attack experiment for these three datasets. In this experiment, each attack method could only change one edge of the original graph to generate adversarial graph. In other words, we set the AME of all attack methods to 1. The set of attacked vertices is the same as the one in the untargeted label attack experiment. We also experiment with direct, indirect, and unlimited attacks, respectively.

The results of single-edge attack are shown in TABLE~\ref{T6}. In general, single-edge attack is a special case of untargeted label attack. Therefore, the results in TABLE~\ref{T6} are consistent with those in TABLE~\ref{T3}. For the unlimited and direct cases, Graphfool still outperforms most of the other attack methods.
FGA is the closest algorithm, with 2\% to 5\% lower ASR on average. Moreover, for Cora and Citeseer, both unlimited Graphfool and direct Graphfool achieve approximately 50\% ASR, but in Pol.blogs, which is much denser than the other two graphs, these two methods only get 18.16\% ASR.

In addition, the ASRs of Graphfool, FGA, and NETTECK in GCN model are higher than those in other node classification algorithms. This is because of the following reasons. First, Graphfool, FGA, and NETTECK are GCN-based graph attack methods. Second, for other node classification algorithms, they all have certain randomness. Single-edge attack only changes one edge in original graph. It may affect more significantly during random process, thus reducing the ASR for these algorithms.

\begin{table*}[!t]
\centering
\caption{Single-edge attack on three datasets.}
\begin{tabular}{c|c|ccc|ccccc}
\hline
\hline
\multirow{3}{*}{Datasets}  & \multirow{3}{*}{Model} & \multicolumn{8}{c}{ASR(\%)}                                                   \\ \cline{3-10}
                           &                        & \multicolumn{3}{c|}{Graphfool} & \multicolumn{5}{c}{BASELINE}                 \\ \cline{3-10}
                           &                        & unlimited  & direct & indirect & FGA   & NATTECK & RL-S2V & GradArgmax & DICE  \\ \hline
\multirow{7}{*}{Cora}      & GCN                    & 70.63      & \textbf{71.71}  & 39.35    & 65.84 & 68.78   & 20.38  & 17.36      & 10.58 \\
                           & GraRep                 & \textbf{43.26}      & 43.26  & 5.08     & 38.48 & 32.35   & 18.86  & 16.27      & 8.08  \\
                           & Deepwalk               & \textbf{45.71}      & 45.71  & 4.27     & 41.54 & 29.36   & 15.42  & 14.06      & 10.58 \\
                           & node2vec               & \textbf{43.26}      & 43.26  & 4.27     & 42.73 & 31.54   & 16.03  & 14.06      & 10.58 \\
                           & LINE                   & 42.73      & \textbf{45.71}  & 5.08     & 42.73 & 28.45   & 17.65  & 16.27      & 9.27  \\
                           & GraphGAN               & \textbf{44.56}      & 43.26  & 3.58     & 39.67 & 26.76   & 16.03  & 15.75      & 10.58 \\ \cline{2-10}
                           & Average                & 48.36      & \textbf{48.82}  & 10.27    & 45.17 & 36.21   & 17.40  & 15.63      & 9.95  \\ \hline
\multirow{7}{*}{Citeseer}  & GCN                    & \textbf{85.74}      & 84.62  & 39.33    & 75.77 & 66.37   & 29.62  & 23.58      & 12.46 \\
                           & GraRep                 & \textbf{49.56}      & 49.56  & 7.23     & 45.32 & 36.53   & 25.76  & 21.57      & 10.37 \\
                           & Deepwalk               & \textbf{47.32}      & 47.32  & 6.59     & 42.58 & 34.87   & 22.43  & 20.64      & 12.46 \\
                           & node2vec               & \textbf{46.59}      & 46.59  & 5.48     & 44.76 & 32.64   & 21.57  & 23.58      & 10.37 \\
                           & LINE                   & 48.23      & \textbf{49.56}  & 5.93     & 43.41 & 35.29   & 26.38  & 24.61      & 11.48 \\
                           & GraphGAN               & \textbf{46.59}      & 46.59  & 6.59     & 42.58 & 34.02   & 23.58  & 21.57      & 9.75  \\ \cline{2-10}
                           & Average                & 54.01      & \textbf{54.04}  & 11.86    & 49.07 & 39.95   & 24.89  & 22.59      & 11.15 \\ \hline
\multirow{7}{*}{Pol.blogs} & GCN                    & \textbf{47.05}      & 47.05  & 12.43    & 44.91 & 33.61   & 6.53   & 7.20       & 5.80  \\
                           & GraRep                 & 8.96       & 8.96   & 2.50     & \textbf{9.89}  & 6.53    & 7.20   & 5.80       & 4.06  \\
                           & Deepwalk               & \textbf{10.53}      & 10.53  & 0.00     & 9.89  & 8.96    & 5.80   & 0.00       & 6.53  \\
                           & node2vec               & 12.42      & 12.42  & 0.00     & \textbf{13.16} & 12.42   & 3.81   & 0.00       & 5.80  \\
                           & LINE                   & \textbf{15.21}      & 15.21  & 2.50     & 8.96  & 13.16   & 0.00   & 3.81       & 2.50  \\
                           & GraphGAN               & \textbf{14.81}      & 14.81  & 0.00     & 11.72 & 14.02   & 3.81   & 0.00       & 2.50  \\ \cline{2-10}
                           & Average                & \textbf{18.16}      & 18.16  & 2.91     & 16.42 & 14.78   & 4.53   & 2.80       & 4.53  \\ \hline
\hline
\end{tabular}
\label{T6}
\end{table*}

\begin{table}[]
\centering
\caption{Attacked vertices' average ratio of different neighbor orders(\%).}
\begin{tabular}{c|ccccc}
\hline
\hline
\multirow{2}{*}{Datasets} & \multicolumn{5}{c}{Neighbor order}   \\ \cline{2-6}
                          & 1    & 2     & 3     & 4     & 5     \\ \hline
Cora                      & 0.12 & 0.36  & 1.79  & 5.22  & 14.22 \\
Citeseer                  & 0.01 & 0.21  & 0.59  & 1.48  & 2.87  \\
Pol.blogs                 & 1.41 & 29.46 & 66.38 & 80.20 & 81.88 \\ \hline
\hline
\end{tabular}
\label{T7}
\end{table}

\begin{table}[]
\centering
\caption{Disturbance-limited attack on Cora.}
\begin{tabular}{c|c|ccccc}
\hline
\hline
\multirow{2}{*}{Metris}  & \multirow{2}{*}{Model} & \multicolumn{5}{c}{Neighbor order}    \\ \cline{3-7}
                         &                        & 1     & 2     & 3     & 4     & 5     \\ \hline
\multirow{7}{*}{ASR(\%)} & GCN                    & 8.14  & 17.86 & 48.14 & 63.75 & 72.41 \\
                         & GraRep                 & 3.33  & 12.50 & 44.17 & 55.84 & 66.67 \\
                         & DeepWalk               & 6.67  & 14.39 & 47.37 & 58.61 & 70.37 \\
                         & Node2vec               & 5.83  & 13.33 & 46.67 & 56.67 & 65.83 \\
                         & LINE                   & 7.50  & 15.01 & 49.17 & 52.17 & 63.33 \\
                         & GraphGAN               & 5.83  & 15.83 & 42.50 & 53.33 & 67.50 \\ \cline{2-7}
                         & Average                & 6.22  & 14.82 & 46.34 & 56.73 & 67.69 \\ \hline
\multirow{7}{*}{AME}     & GCN                    & 18.73 & 16.86 & 12.63 & 8.41  & 5.32  \\
                         & GraRep                 & 19.76 & 17.93 & 15.04 & 8.57  & 6.48  \\
                         & DeepWalk               & 19.23 & 18.69 & 14.39 & 9.01  & 6.21  \\
                         & Node2vec               & 19.54 & 18.63 & 13.58 & 9.25  & 7.01  \\
                         & LINE                   & 19.06 & 17.59 & 14.69 & 9.46  & 7.74  \\
                         & GraphGAN               & 19.38 & 18.06 & 15.43 & 10.39 & 6.87  \\ \cline{2-7}
                         & Average                & 19.28 & 17.96 & 14.29 & 9.18  & 6.61  \\ \hline
\hline
\end{tabular}
\label{T8}
\end{table}

\begin{table}[]
\centering
\caption{Disturbance-limited attack on Citeseer.}
\begin{tabular}{c|c|ccccc}
\hline
\hline
\multirow{2}{*}{Metris}  & \multirow{2}{*}{Model} & \multicolumn{5}{c}{Neighbor order}    \\ \cline{3-7}
                         &                        & 1     & 2     & 3     & 4     & 5     \\ \hline
\multirow{7}{*}{ASR(\%)} & GCN                    & 3.96  & 17.82 & 27.66 & 44.55 & 52.50 \\
                         & GraRep                 & 1.98  & 14.85 & 23.76 & 39.60 & 46.53 \\
                         & DeepWalk               & 2.97  & 12.87 & 21.78 & 40.59 & 47.52 \\
                         & Node2vec               & 3.96  & 15.84 & 19.80 & 36.63 & 47.52 \\
                         & LINE                   & 2.97  & 11.88 & 22.77 & 41.58 & 51.48 \\
                         & GraphGAN               & 2.97  & 13.56 & 22.77 & 37.62 & 49.50 \\ \cline{2-7}
                         & Average                & 3.14  & 14.47 & 23.09 & 40.10 & 49.18 \\ \hline
\multirow{7}{*}{AME}     & GCN                    & 19.26 & 17.96 & 14.87 & 9.87  & 7.95  \\
                         & GraRep                 & 19.94 & 18.79 & 16.04 & 10.93 & 10.01 \\
                         & DeepWalk               & 19.76 & 19.04 & 16.88 & 11.57 & 9.36  \\
                         & Node2vec               & 19.53 & 18.54 & 17.06 & 12.03 & 9.51  \\
                         & LINE                   & 19.87 & 19.02 & 15.96 & 11.42 & 8.54  \\
                         & GraphGAN               & 19.87 & 18.33 & 15.83 & 12.21 & 8.87  \\ \cline{2-7}
                         & Average                & 19.71 & 18.61 & 16.11 & 11.34 & 9.04  \\ \hline
\hline
\end{tabular}
\label{T9}
\end{table}

\begin{table}[]
\centering
\caption{Disturbance-limited attack on Pol.blogs.}
\begin{tabular}{c|c|ccccc}
\hline
\hline
\multirow{2}{*}{Metris}  & \multirow{2}{*}{Model} & \multicolumn{5}{c}{Neighbor order}    \\ \cline{3-7}
                         &                        & 1     & 2     & 3     & 4     & 5     \\ \hline
\multirow{7}{*}{ASR(\%)} & GCN                    & 10.64 & 74.47 & 89.36 & 91.30 & 93.30 \\
                         & GraRep                 & 8.53  & 72.25 & 85.36 & 88.36 & 91.30 \\
                         & DeepWalk               & 7.50  & 67.53 & 86.27 & 87.87 & 89.47 \\
                         & Node2vec               & 5.63  & 65.48 & 84.26 & 86.49 & 87.87 \\
                         & LINE                   & 7.50  & 68.83 & 85.25 & 87.87 & 87.87 \\
                         & GraphGAN               & 4.87  & 62.57 & 82.03 & 86.49 & 89.47 \\ \cline{2-7}
                         & Average                & 7.45  & 68.52 & 85.42 & 88.06 & 89.88 \\ \hline
\multirow{7}{*}{AME}     & GCN                    & 15.85 & 10.68 & 6.38  & 5.84  & 5.74  \\
                         & GraRep                 & 17.53 & 12.84 & 9.07  & 6.98  & 6.73  \\
                         & DeepWalk               & 16.45 & 12.25 & 8.83  & 7.59  & 7.53  \\
                         & Node2vec               & 16.88 & 13.47 & 8.52  & 6.26  & 6.37  \\
                         & LINE                   & 16.27 & 13.09 & 8.04  & 7.23  & 6.81  \\
                         & GraphGAN               & 17.26 & 13.62 & 9.21  & 8.14  & 7.15  \\ \cline{2-7}
                         & Average                & 16.71 & 12.66 & 8.34  & 7.01  & 6.72  \\ \hline
\hline
\end{tabular}
\label{T10}
\end{table}

\subsubsection{Disturbance-limited attack}

To improve the attack concealment, one may want to limit the perturbation to a certain range. This is called disturbance-limited attack. In disturbance-limited attack, attackers only change the edges of the subgraph which is composed by the attacked vertex and its neighbors. For each attacked vertex, we first calculate its $k$-order neighbors (vertices whose link distance to the attacked vertex are less than $k$) and construct the subgraph accordingly. Then we perform graph attack in this subgraph using unlimited Graphfool attack method. In this experiment, we set the range of $k$ from 1 to 5.
The value $k$ is corresponding to the size of the modifiable subgraph. To quantify the size of these subgraphs in the original graph, for each dataset, we calculate the average ratio of the number of vertices in the  $k$-th order subgraph of each attacked vertex to the corresponding number of vertices in original graph. The results are shown in TABLE~\ref{T7}.

TABLE~\ref{T8}, TABLE~\ref{T9}, and TABLE~\ref{T10} show the results of disturbance-limited attack on three datasets. With the increase of neighbor order, ASR in every node classification algorithm increase significantly, while AME is monotonously decreasing. This is consistent with the general idea that when the size of the constructed subgraph gets larger, the Graphfool attack is more likely to succeed. Moreover, for the sparse dataset Cora and Citeseer, when the neighbor order of each attacked vertex is 5, the average sizes of subgraphs are only 14.22\% and 2.87\%. However, Graphfool attack also could get 67.69\% and 49.18\% average ASR, 6.61 and 9.04 average AME, respectively. This implies that even if we limit the disturbance to a small local subgraph of the attacked vertex, we can still perform effective attacks. However, For the dense graph Pol.blogs, when the neighbor order reaches 5, the average size of subgraphs becomes 81.88\%. It covers most of the original graph and the results are close to the unlimited case in TABLE~\ref{T3}.

\section{Conclusion\label{Conclusion}}
In this paper, we propose Graphfool, a graph attack technique to mislead the node classification algorithms. This method uses two-layer GCN model as the attack model, constructs the decision boundary according to the classification results, and implements attack by modifying edge relationship of adjacency matrix. The experiments suggest that Graphfool can generate smaller perturbation and obtain the higher attack success rate simultaneously, which demonstrate the vulnerability of current graph embedding algorithms.

\bibliographystyle{IEEEtran}
\bibliography{refer}

\end{document}